\documentclass[%
    pdflatex, sn-mathphys]{sn-jnl}
\usepackage[utf8]{inputenc}
\usepackage{amsmath} %ambiente matematico generico
\usepackage{bm} %ambiente matematico generico
\usepackage{amssymb} %simboli matematici
\usepackage{amsthm}  %teoremi e simili
\usepackage{centernot} 
\usepackage{wasysym}
\usepackage{bbold}
\newcommand{\indep}{\rotatebox[origin=c]{90}{$\models$}}
\usepackage{subcaption}

\theoremstyle{plain}

\theoremstyle{plain}
\newtheorem*{theorem*}{Theorem}
\theoremstyle{plain}
\newtheorem{theorem}{Theorem}[section]
\theoremstyle{remark}

\theoremstyle{remark}

\theoremstyle{plain}
\newtheorem{corollary}{Corollary}[section]
\theoremstyle{definition}

\theoremstyle{definition}

\theoremstyle{plain}

\theoremstyle{remark}
\newtheorem{example}{Example}[section]

\jyear{2023}%

\newcommand{\EE}{\mathbb{E}}
\newcommand{\RR}{\mathbb{R}}

\begin{document}

\title[How to be fair?]{How to be fair? A study of label and selection bias.}

% The right type of fairness for the right type of bias
% Selecting fairness intervention based on the kind of bias
% Different Biases require different interventions
% When are fair interventions fair? ...
% Selecting the right type of fairness...
% Being fair is fair-ly difficult

\author*[1]{\fnm{Marco} \sur{Favier}}\email{marco.favier@uantwerpen.be}

\author[1]{\fnm{Toon} \sur{Calders}}

\author[1]{\fnm{Sam} \sur{Pinxteren}}

\author[1]{\fnm{Jonathan} \sur{Meyer}}

\affil[1]{\orgname{University of Antwerp}, \orgaddress{\city{Antwerp}, \country{Belgium}}}

\abstract{%
It is widely accepted that biased data leads to biased and thus potentially unfair models. Therefore, several measures for bias in data and model predictions have been proposed, as well as bias mitigation techniques whose aim is to learn models that are fair by design. Despite the myriad of mitigation techniques developed in the past decade, however, it is still poorly understood under what circumstances which methods work. Recently, Wick et al. showed, with experiments on synthetic data, that there exist situations in which bias mitigation techniques lead to more accurate models when measured on unbiased data. Nevertheless, in the absence of a thorough mathematical analysis, it remains unclear which techniques are effective under what circumstances. 
We propose to address this problem by establishing relationships between the type of bias and the effectiveness of a mitigation technique, where we categorize the mitigation techniques by the bias measure they optimize. In this paper we illustrate this principle for label and selection bias on the one hand, and demographic parity and ``We're All Equal'' on the other hand. Our theoretical analysis allows to explain the results of Wick et al. and we also show that there are situations where minimizing fairness measures does not result in the fairest possible distribution.
} 


\maketitle

% SyRI
\section{Introduction}
% It is widely accepted that biased data leads to biased and thus potentially unfair models.

%In the past 10 years, since the first papers on the topic began to emerge, many different perspectives have been provided on the concept of fairness in machine learning and how to approach the problem. Following this prolific decade, a plethora of different measures have been proposed to calculate how unfair a model is, and similarly, a multitude of different methods have been developed in parallel to minimize these measures. The, then hypothetical, problem has proven to be so relevant to reality that it has exceeded the strictly academic sphere on more than one occasion, spilling over into common media.[insert citations]

In numerous cases it was shown that models trained on biased data may exhibit undesirable behavior towards certain groups in the population in a systematic way. For instance, according to Obermeyer et al.~\cite{obermeyer_dissecting_2019}, there was racial bias in a widely used algorithm in health care in the US, such that black patients assigned the same level of risk by the algorithm were on average more sick than white patients. Obermeyer et al. argue that this bias was the result of health \emph{costs} being used as a proxy for health \emph{needs} while historically less money is spent on black patients for the same needs. The type of bias in this case is called \emph{label bias}, indicating that the labels (health cost) do not properly reflect the prediction target (health status).

The Dutch childcare benefits scandal, also known as the \emph{toeslagenaffaire}~\cite{Belastingdienst}, is another significant example of the potential consequences of implementing systems to prevent fraud without proper consideration of potential biases. In an effort to minimize the risk of fraud, the Dutch tax administration implemented a system to select child benefit recipients for audits. However, this system resulted in thousands of parents being falsely accused of fraudulently claiming benefits and being required to return the benefits they had received. Additionally, it was found that the system disproportionately affected individuals with non-Dutch nationality, as the model used by the administration considered them to have a higher risk of tax fraud than Dutch nationals. An analysis of the dataset used to train the model, the \emph{Fraude Signalering Voorziening}, revealed that it heavily relied on denunciations and tips from citizens~\cite{Belastingdienst}. This reliance on denunciations raises the possibility of bias, as it is plausible to assume that people with a different nationality than Dutch were more likely to be reported anonymously by Dutch citizens when committing a crime than Dutch people committing the same crime. This kind of bias can be identify as \emph{selection bias}: a situation where a sample of data is not descriptive of the population it is intended to represent due to an under or over representation of certain groups. This is a possibility in the toeslagenaffaire, where fraudulent non-Dutch residents may have been disproportionately sampled while fraudulent Dutch ones were overlooked.

% Therefore, several measures for bias in data and model predictions have been proposed, as well as bias mitigation techniques whose aim is to learn models that are fair by design.
In order to identify bias in data and model predictions, several measures have been proposed. One example of such a measure is \emph{demographic parity difference} (DPD) which measures the difference in probability of getting assigned the positive label between two predefined groups: a legally protected group and its complement. See Barocas et al.~\cite[Chapter 3]{fairnessbook} for a systematic overview of the most common fairness measures, connecting them to the fairness criteria. For instance, DPD measures to what degree a dataset satisfies the fairness criterion \emph{independence} which states that label and sensitive group membership should be independent. Complementary to the bias measures, \emph{fairness interventions} were proposed to learn models that are fair by design, in the sense that they are constrained to produce models that obey a bias measure of choice. The typical workflow of a fair machine learning practitioner can hence be divided into the following three steps: (1) selecting a fairness metric, (2) minimizing that metric for a model, and (3) claiming that the model is fair. For an overview of bias mitigation techniques we refer to the \emph{Fairness Library\footnote{\url{https://axa-rev-research.github.io/fairness-compass/src/main/library/}}}~\cite{ruf2022tool}.

% Despite the myriad of mitigation techniques developed in the past decade, however, it is still poorly understood under what circumstances which methods work.
Despite the large number of fairness measures and interventions, however, it is still poorly understood what the exact effects of the different fairness interventions are and in which situations they should be applied. Consider for instance the Healthcare example with label bias and the toeslagenaffaire example with selection bias. Is there any bias mitigation technique we could apply on the biased data such that the resulting model would be fair? The current state of the fairness research field does not allow us to answer this question. Moreover, it is also not immediately clear which definition of fairness would apply in these examples.

Earlier works considered fairness measures as constraints that should be legally enforced~\cite{feldman2015certifying}. We call this the \emph{legal constraint framework}. In this framework it makes sense to maintain accuracy at a high level while constraining models to those that satisfy the fairness constraint. In such a setting, inevitably there is a tension between the degree of fairness on the one hand, and the accuracy of models on the other, called the \emph{fairness-accuracy trade-off}. Works by legal scholars, such as~\cite{wachter2021fairness}, however, have shown this approach to be inadequate because of the impossibility to have a strict mathematical interpretation of the legal definition of discrimination. 

\begin{figure}[tbh]
\includegraphics[width=\columnwidth]{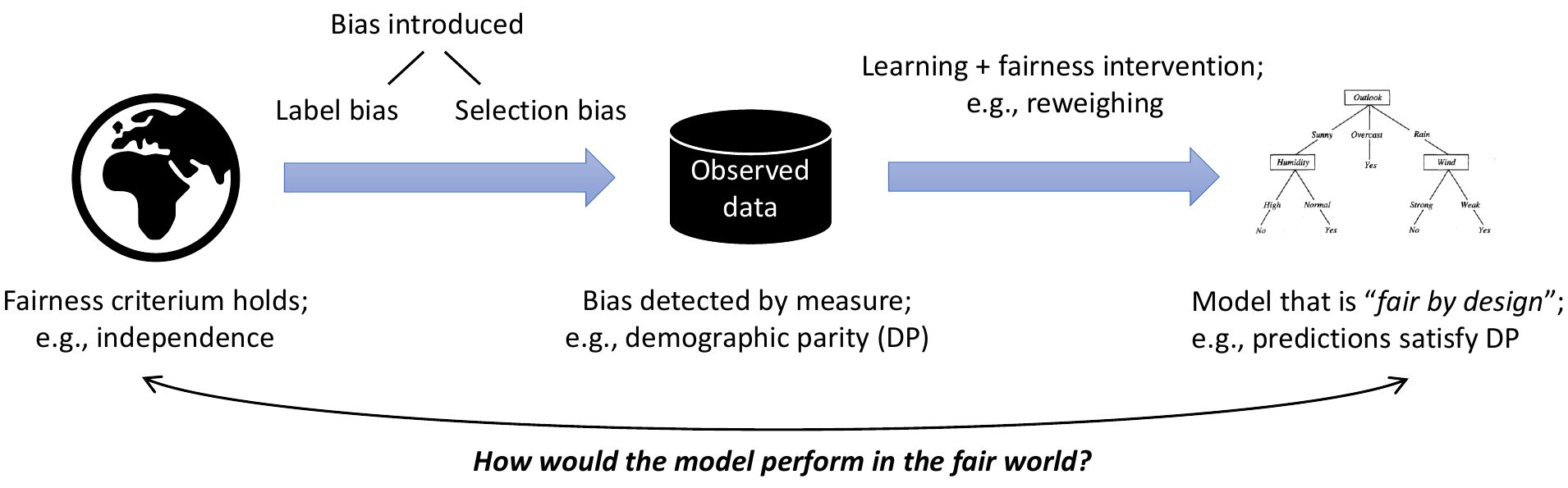}
\caption{The Fair World Framework on which out theory is based. We assume a fair world in which a fairness criterion holds. The observed data, however, is obtained through a biased process. Fairness-aware techniques learn models in the biased data while simultaneously optimizing a fairness measure. This results in a model that is ``fair by design.''}
\label{fig:framework}
\end{figure}

Therefore, in this paper we consider the alternative framework called \emph{Fair World Framework} depicted in Figure~\ref{fig:framework}. We assume a fair, underlying world, which we can only observe through a biased dataset. A good bias mitigation technique would then allow to infer, from the biased data, a model that would perform accurately in the fair world. In this setting, the choice of the right mitigation strategy will depend on the type of bias introduced, and on assumptions we can make about the fair world. Notice that this setting is more challenging than the legal constraint framework; it is insufficient to optimize for the measure that corresponds to the fairness criterion satisfied in the real world. For instance, in the Healthcare example we may assume that in the ``fair world,'' the label given to people corresponds to their true health status. By using Health cost instead, a label bias was introduced. Undoing the label bias is more complex than rebalancing the labels by constraining the demographic parity difference measure. Instead, we need to make sure the labels of the right patients get changed as otherwise we may stray away from the fair world even further. Moreover, recently, Wick et al.~\cite{wick2019unlocking} showed, with experiments on synthetic data, that there exist situations in which bias mitigation techniques lead to more accurate models when measured on unbiased data. Their experiment fits well with our framework, as they construct a fair dataset that satisfies independence, then introduce label and selection bias, and subsequently learn a model while employing a fairness intervention on the biased data. This model is then shown to perform with higher accuracy on the unbiased data than a model learned on the biased data without fairness intervention. 
% Nevertheless, in the absence of a thorough mathematical analysis, these results are anecdotal and it remains unclear which techniques are effective under what circumstances.

Nevertheless, a thorough mathematical analysis connecting which techniques are effective under what circumstances is still needed, the lack of which may dampen the importance of these results that risk to be judged as simply anecdotal.

We propose to address this problem by establishing relationships between the type of bias and the effectiveness of a mitigation technique. The contributions of this paper are as follows:
\begin{enumerate}
\item We introduce the fair world framework, and formally define two bias introduction processes: label bias and selection bias. We further consider two fairness criteria that can be satisfied in the fair world. The choice of bias process and fairness criterion will be assumptions we have to make when using the framework. 
\item For all four combinations of a fairness criterion and a bias introduction process, we show how it transforms the data resulting in sets of properties that need to be satisfied by the biased dataset. These properties allow us to check consistency of our assumptions.
% \item We connect fairness interventions to the bias measure they optimize and hence the fairness criterion they implicitly assume.
\item We show for each combination of fairness criterion and bias introduction process whether a fairness intervention aiming for that same criterion on the biased dataset allows for finding the fair model. These results allow us to explain theoretically the results obtained by the simulation of recent papers.
\end{enumerate}
The relevance and impact of our work is two-fold. On the one hand, it opens up a new way of addressing fairness. By framing fairness as accuracy in the underlying fair world, practitioners can select fairness interventions based on assumptions about the process that introduced bias in the unfair world and the fairness criteria that hold in the fair world. We believe such assumptions to be more intuitive and natural to make. On the other hand, the theoretical results allow for explaining the reasons behind important recent empirical findings.

\section{Related Work}
\subsection{Measuring fairness.} In the algorithmic bias literature, several measures of fairness of models and data have been proposed, including group fairness measures; such as statistical parity, equal opportunity, calibration, individual fairness measures, and causality-based measures. These measures concentrate on different aspects of the classifier: disparity of impact, differences in error rates, correlation between a sensitive attribute and predicted label. We refer to surveys like \cite{mehrabi2021survey} or \cite{fairnessbook} for an overview of different measures. Although for each of these measures convincing motivating examples exist, unfortunately it is not possible to combine them in a meaningful way~\cite{kleinberg2016inherent}. 

This abundance of measures can be confusing and frustrating for practitioners. Impossibility results are showing that fairness criteria that seem mandatory from an ethical perspective, are internally inconsistent, discouraging data scientists as they imply that any method, no matter how carefully applied, will break at least one fairness criterion and as such will be subject to criticism. To address this issue, researchers have come up with categorizations of which fairness measure is most suitable for a given situation \cite{ruf2021towards}, and visual tools to interactively explore fairness of data and models such as FairSight \cite{ahn2019fairsight} and FairVis \cite{cabrera2019fairvis} have been proposed.
\subsection{Bias mitigation techniques.} Plenty of model inference techniques that are \emph{fair by design} have been developed, many of which are included in tools like AIF 360~\cite{bellamy2019ai}. Models are fair by design in the sense that they optimize a particular measure; models output by these inference techniques score by design favourably for that particular measure. It is, however, unclear to what extent satisfying a measure by design makes a classifier fair. %For instance, suppose that a classifier to select CVs picks up a relationship between gender and profession, and exploits this in its predictions. Such a classifier would fail statistical parity measures. A post-processing technique to correct the bias of this classifier could change its  predictions to no longer correlate with gender. There are, however, different ways to correct the classifier; it could upgrade negative labels to positive ones for random members of a disadvantaged group, or randomly downgrade members of a privileged group. As such, where a poor measure of fairness indicates an issue with a classifier, a good value for a fairness measure does not necessarily mean that a classifier is more fair. 
That is, blindly improving a fairness measure may not necessarily lead to an ethical solution; we need a deeper understanding of bias in data and models to better guide bias mitigation techniques.

In addition, interventions that improve fairness metrics tend to decrease the accuracy \cite{menon2018cost,corbett2017algorithmic, chen2018my,cooper2021emergent}. This phenomenon is known as the fairness-accuracy trade-off. However, as noticed by other authors \cite{dutta2020there,wick2019unlocking}, this trade-off is \emph{in se} a false one because fairness and accuracy are measured with respect to a dataset we assume is biased. As a result our measurements of accuracy and fairness will be biased as well. Controlled experiments with datasets containing both biased and unbiased labels~\cite{lenders2023real} or simulated data~\cite{wick2019unlocking} confirm this claim.
\subsection{Understanding bias and its relation to fairness.}
Recently, the study of algorithmic fairness seems to be moving from fairness as just a constraint to satisfy with various techniques (like those in \cite{bellamy2019ai}) to fairness as data fallacy and bias to overcome. In that regard the study of fairness under the causal framework is especially rich~\cite{kusner2017counterfactual, loftus2018causal}.
It has been shown that fairness cannot be well assessed merely based on correlation or association as the aggregate relationship between the sensitive attribute and the output variable may disappear or reverse when accounting for other relevant factors \cite{ghai2022d}. Still causal fairness often requires the a priori knowledge of a causal graph and some classical techniques on Bayesian networks like Expectation-Maximization may fail \cite{calders2010three}. 

A particularly inspiring work for this paper comes from \cite{friedler2016possibility,friedler2021possibility}, in their work Friedler et al. assume the existence of three different spaces: the construct, observed and decision spaces. Their definition of fairness depends on how the three world interact with each other, and in particular how information from the construct space gets modified and distorted in the decision space. We use this conceptual framework as well and we expand it further similarly defining bias as possible interaction between these spaces.

%The existence of the accuracy-fairness trade-off phenomenon has being proven and shown by different scholars \cite{menon2018cost,corbett2017algorithmic, chen2018my} and then challenged by others \cite{dutta2020there,wick2019unlocking}, 
%our work aligns with the latter group, in particular further developing the results from \cite{wick2019unlocking}. The study of algorithmic fairness seems to be moving from fairness as just a constraint to satisfy with various techniques (like those in \cite{bellamy2019ai}) to fairness as data fallacy and bias to overcome. In that regard the study of fairness under the causal framework is especially rich.  \cite{kusner2017counterfactual, loftus2018causal}, still causal fairness often requires the a priori knowledge of a causal graph and some classical techniques on Bayesian networks like Expectation-Maximization may fail \cite{calders2010three}.\\

\section{Notation}
In this paper, we will discuss the difference between the distribution of fair data and the distribution of data available in dataset $D$. The main variables at our disposal are:

\begin{itemize}
    \item $X = [X_1,\dots, X_n]$, a random vector containing the attributes $X_i$.
    \item $A\in \{a_0,a_1\}$, a random variable that represents the sensitive attribute. For convenience we consider it a binary variable, with $A=a_0$ representing the deprived population and $A=a_1$ representing the privileged population.
    \item $Y\in\{y_0,y_1\}$, a binary label that we are interested in predicting for each datapoint. We consider $Y=y_0$ to be the unfavorable outcome and $Y= y_1$ to be the favorable one.
\end{itemize}
As a result of using an inaccessible fair distribution in contrast to the observed one, we need to use different notation for the two distributions:
\begin{itemize}
    \item $P(Y,X,A)$ is the original, fair distribution of the data, which we do not have access to, but which we would like to predict.
    \item $P_D(Y,X,A)$ is the distribution from which dataset $D$ has been sampled.
\end{itemize}
It's important to keep in mind that we cannot guarantee that the two distributions will be the same. In fact, in general, we will have
\[
 P_D(Y,X,A)\neq P(Y,X,A)
\]
For convenience, we will write $P(y,x,a)$ instead of $P(Y=y,X=x,A=a)$ when this does not cause confusion. The same rule applies to $P_D$.\\ Also, to avoid needless repetitions, the reader should be aware that every equation where probabilities are shown holds for all $x\in X$, unless stated otherwise.

As already mentioned, the use of a binary sensitive attribute is out of convenience: for non-binary attributes all the results still hold by substituting $a_0,a_1$ with $a_i,a_j$ whenever necessary. However, the same cannot be said about the label, which needs to be binary for many of the propositions to be valid.

\section{Fair World Framework}
In our work, we assume the existence of a fair world that is inaccessible to us and in which fairness is satisfied by default. The distribution, $P(y,x,a)$, from this world is distorted by a biasing procedure, resulting in a misrepresentation of the world, $P_D(y,x,a)$. Our focus will be on two different worldviews where different definitions of fairness can hold.
\subsection{Statistical Parity}
Statistical parity assumes that, across all possible sensitive groups, the distribution of the label is statistically the same. This means that people from different populations have equal access to the label. This concept has also been referred to as ``equal base rates" in the literature \cite{kleinberg2016inherent} when the independence property is satisfied by the original data.
Formally, according to the fair distribution, the following constraint holds:
\[
Y\indep A
\]
In other words, the distribution of the label is independent of the sensitive attribute.\\
This doesn't imply that the groups are perfectly identical. They can have different characteristics that affect their likelihood of achieving the label differently. For example, if the label is ``being a sports enthusiast", and hypothetically men liked football more while women liked volleyball equally, the label would still be equally distributed between the two groups, meeting the criteria for statistical parity.
There are multiple measures used to check if statistical parity holds for a probabilistic model \cite{bellamy2019ai,feldman2015certifying}. We will focus on the \emph{demographic parity difference} (DPD). Given a probabilistic model $P_M(y\vert\, x,a)$, we can calculate the demographic parity difference {DPD}$(P_M)$  as follows:
\[
    \left\lvert
    \sum_{(y,x,a)\in D }\left[\frac
    {P_M(y_1\vert\,  x,a_0)\mathbb{1}_{a_0}(a)}
    {\vert\{(y,x,a)\in D\colon a = a_0 \} \vert}
    -
    \frac
    {P_M(y_1\vert\,  x,a_1)\mathbb{1}_{a_1}(a)}
    {\vert\{(y,x,a)\in D\colon a = a_1 \} \vert}
    \right]\right\rvert
\]
We can see that, as the size of the dataset grows, assuming $P_M(y\vert\,  x,a) \to P_D(y\vert\,  x,a)$ the difference converges to
\[
 \text{DPD}(P_M) \xrightarrow{\lvert D \rvert \to \infty} \left\lvert P_D(y_1\vert\,  a_1) - P_D(y_1\vert\,  a_0) \right\rvert 
\]

\subsection{We're All Equal}
The fairness assumption we will present in this section is what we call ``We're All Equal" (WAE). According to this worldview, every person has an equal probability of receiving the positive label regardless of their sensitive attribute. For example, two candidates for a job with the same past experiences should have an equal chance of getting hired, regardless of their race. This does not mean that the set of features is independent from the sensitive variable, but rather that the sensitive attribute does not matter for the choice of the label when the rest of the attributes are known. This also means that it is considered acceptable to observe some demographic disparity in our data if some positive traits for the label are correlated with the sensitive attribute%
% , especially when this correlation is not causal
. This can be formalized with the following independence constraint:
\[
    Y\indep A\vert\,  X
\]
Both WAE and statistical parity are two extremes of a more general definition: conditional statistical parity. This definition assumes the existence of a subset $S\subseteq X$ of variables such that $Y\indep A\, \vert\, S$. Statistical parity is the case where $S=\emptyset$, while WAE is the case where $S=X$.

\section{Bias}
Fully comprehending the reasons why bias exists in our society, its reflection in datasets, and how to address its effects, is a problem deeply rooted in sociology, philosophy, and history, with extremely complex causes and consequences. Computer science alone cannot solve this problem.
However, to mathematically study how bias affects data, we adopt what we call a ``banality of evil" principle: we assume that bias primarily depends on the sensitive attribute, which is a simplification necessary for the mathematical analysis of its effect. Nonetheless, we believe that the results we found provide valuable insights even in more general cases.

We will now proceed to discuss and formalize different types of biases, focusing on two types: label bias and selection bias.
\subsection{Label Bias}
Label bias occurs when individuals from different sensitive groups are treated differently solely based on their sensitive feature. This ranges from direct discrimination against minorities to unjustified privileges of some elitist groups. Excluding the label itself, the representation of each individual in the dataset is not affected by the biasing procedure. 

Formally, we will assume the existence of a new latent binary variable $C\in \{c_0,c_1\}$ that indicates whether the datapoint in our possession has been influenced by label bias or not. If the candidate has undergone label bias $(C=c_1)$, the label is changed, while if the candidate hasn't been affected by the bias $(C=c_0)$, no changes occur. We will also assume that $C$ depends only on the original label and the sensitive attribute. Formally,
\[
C\indep X\mid A, Y
\]
It should be noted that since C directly depends on the original label and the sensitive attribute, it can be chosen to either discriminate, by setting a high probability of change for minorities with the positive label, or to favor the already privileged group by randomly upgrading the label to the positive one, or even to do both. We can already demonstrate some early results on how label bias affects the distribution of the fair world.

\begin{theorem}\label{teo:label bias}
Under label bias, the relation between the conditional distributions $P(y_1\mid x, a)$ and $P_D(y_1\mid x, a)$ is linear, that is
\[
     \alpha_a P(y_1\mid x, a) + \beta_a P_D(y_1\mid x, a)+\gamma_a = 0 \quad 
\]
for some $(\alpha_a,\beta_a, \gamma_a) \in \RR^3\smallsetminus\{(0,0,0)\}$, while $P_D(x,a)= P(x,a)$
\begin{proof}
It is easy to see that the only variable affected is $Y$ and it follows this relation\\
\begin{equation}
\label{eq:discrimination}
    P_D(y_1\mid x,a) = P(c_0\mid y_1, a)P(y_1\mid x,a) + P(c_1\mid y_0, a)P(y_0\mid x,a)    
\end{equation}\\
that is because the observed value of $y$ in the dataset is either the true one, if there has been no change of label $(C=c_0)$, or the opposite if $C=c_1$.
Because $Y$ is binary we can rewrite the above equation as\\
\[
    \alpha_a P(y_1\mid x, a) + \beta_a P_D(y_1\mid x, a)+\gamma_a = 0
\]\\
where $\alpha_a:= P(c_0\mid a,y_1)- P(c_1\mid a,  y_0)$, $\beta_a:=-1$ and $\gamma_a:= P(c_1\mid a,  y_0)$.
\end{proof}
\end{theorem}

\subsection{Selection Bias}
Selection bias occurs when the selection of data points is not representative of the underlying distribution. This can occur if the mechanism used to collect data points is flawed. To model the selection bias we once again use a latent variable $K\in\{k_0,k_1\}$ which symbolize when a data point is kept in our dataset $(K=k_1)$ or not kept $(K=k_0)$. So an element $(y,x,a,k)$ is visible and belongs to $D$ if and only if $K=k_1$.  As with the previous bias notion, selection bias can either disadvantage the deprived group, advantage the privileged one, or both. We will further assume that the effect of the variable $X$ on $K$ is negligible. So formally, even if the process differs from the label bias case, the same independence constraint holds:
\[
K\indep X\mid A, Y
\]
Similarly as the previous bias case we can show some early properties of the probability distribution under selection bias.

\begin{theorem}\label{teo:sampling+unaware}
    Under selection bias, the conditional probabilities for the unfair world follow these
    relationships
    \begin{align*}
        \frac{P_D(y_1\mid x, a)}{P_D(y_0\mid x, a)} &=  \delta_a\cdot \frac{P(y_1\mid x, a)}{P(y_0\mid x, a)} \\
        \frac{P_D(y_1\mid  a)}{P_D(y_0\mid a)} &=  \delta_a\cdot \frac{P(y_1\mid a)}{P(y_0\mid a)} 
    \end{align*}
    where $\delta_a := P(k_1\mid y_1,a)/P(k_1\mid y_0,a)$.
    \begin{proof}
    Since a datapoint is present in the database if and only if $K=k_1$ we have
    \[
        P_D(y\mid x, a) = P(y \mid  k_1, x, a)
    \]
    Then by Bayes' theorem
    \[
        P(y \mid  k_1, x, a) = \frac{P(k_1\mid y,a)P(y\mid x,a)}{P(k_1\mid  x, a)} 
    \]
    so
    $$
        \frac{P_D(y_1\mid x, a)}{P_D(y_0\mid x, a)} = \frac{P(k_1\mid y_1,a)}{P(k_1\mid y_0,a)} \frac{P(y_1\mid x,a)}{P(y_0\mid x,a)}
    $$\\
    and similarly the result holds for $P_D(y\mid a)$.
    \end{proof}
    \end{theorem}
\section{Worldview and Bias Combinations}\label{sec:Worldview and Bias}
We now mathematically study how the combination of each fairness worldview and bias process interact with each other. The advantage of this, as previously mentioned, is that finding an accurate model on the fair world corresponds to a model that also satisfies the fairness constraint considered, overcoming the accuracy-fairness trade-off issue. Moreover this modeling results in a mathematically rich framework that can be further explored.

\subsection{Statistical Parity and Label Bias}\label{subsec:parity+label}
The first combination we explore is how a world satisfying statistical parity is impacted by label bias. The first theorem demonstrates that if this occurs, there are conditions that the unfair probability must meet.
\begin{theorem}\label{Teo:stat+label}
Let $P_D$ be a probability distribution resulting from label biasing a distribution satisfying statistical parity. Then the following condition must hold
\[
\bigcap_{a\in A} \left[1-\frac{P_D(y_0\vert\, a)}{\max_{x\in X} P_D (y_0\vert\, x,a)},\frac{P_D(y_1\vert\, a)}{\max_{x\in X} P_D(y_1\vert\, x,a)}\right]\neq\emptyset
\]
    \begin{proof}
    WLOG let's suppose $P(c_1\vert\, y_0,  a)<P(c_0\vert\, y_1, a)$ and consider Equation (\ref{eq:discrimination}). As a consequence we get the following inequalities
    \begin{align*}
        P(c_0\vert\, y_1, a) \geq \max_{x\in X} P_D(y_1\vert\, x,a)\\
        P(c_1\vert\, y_0,  a)\leq \min_{x\in X} P_D(y_1\vert\, x,a)
    \end{align*} 
    Let's now consider $\EE_{x\sim P_D}\left[ P_D(y_1\vert\, x,a)\vert\, A \right]$, because $P_D(x\vert\, a) = P(x\vert\, a)$, Equation (\ref{eq:discrimination}) becomes
        \[
            P_D(y_1\vert\, a) = P(c_0\vert\, y_1, a)P(y_1) +P(c_1\vert\, y_0,  a)P(y_0)
        \]
        therefore we have that
        \begin{align*}
            P_D(y_1\vert\, a)\geq & P(y_1)\max_{x\in X} P_D(y_1\vert\, x,a)\\
            P_D(y_1\vert\, a)\leq & P(y_1)+P(y_0)\min_{x\in X} P_D(y_1\vert\, x,a)
        \end{align*}
        These inequalities can be rewritten as the following
        \[
         1-\frac{P_D(y_0\vert\, a)}{\max P_D(y_0\vert\, x,a)}   \leq P(y_1) \leq \frac{P_D(y_1\vert\, a)}{\max P_D(y_1\vert\, x,a)}
        \]
        Since $P(y_1)$ is independent from $A$, we get the result in the thesis. 
    \end{proof}
\end{theorem}
As a Corollary we have the following
\begin{corollary}\label{coro: example}
Under the conditions of Theorem \ref{Teo:stat+label}, one of the following must happen:
$$\max_{x\in X} P_D (y\vert\, x,a) < 1 \text{ for some } a\in A,y\in Y \text{ or } 
P_D(y_1\vert\, a_0)= P_D(y_1\vert\, a_1) 
$$
\end{corollary}
The Corollary shows that the condition in Theorem \ref{Teo:stat+label} is not always trivially solved by any distribution. It's important to note that the condition shown is, to some extent, also sufficient : if the condition holds it is possible to define fair probability distributions that may have generated the unfair distribution. This is proved in the following theorem:
\begin{theorem}\label{teo: retrive label+DP}
    Let $P(y_1)\in[0,1]$ be an element of the following non-empty set
    \begin{equation}
    \label{eq:interval}
    \bigcap_{a\in A} \left[1-\frac{P_D(y_0\vert\, a)}{\max_{x\in X} P_D (y_0\vert\, x,a)},\frac{P_D(y_1\vert\, a)}{\max_{x\in X} P_D(y_1\vert\, x,a)}\right]
    \end{equation}
    Then the conditional distributions $P(c\vert\, y,a)$ that satisfy the following conditions
    \begin{align*}
        P_D(y_1\vert\, a) =& P(c_0\vert\, y_1, a)P(y_1) +P(c_1\vert\, y_0,  a)(1-P(y_1))\\
        P_D(y_1\vert\, x,a) \in& [P(c_1\vert\, y_0,  a),P(c_0\vert\, y_1, a)]\cup [P(c_0\vert\, y_1,  a),P(c_1\vert\, y_0, a)]
    \end{align*}
    are all and only the distributions that generate $P_D(Y,X,A)$ according to the label bias model under statistical parity. In particular the set of possible $P(c\vert\, a,y)$ is not empty.
    \begin{proof}
        It should be clear from the proof of Theorem \ref{Teo:stat+label} why the conditions for $P(c\,\vert\, a,y)$ must hold when $P_D(Y,X,A)$ follows our model, so let's prove that if the conditions are respected then we can indeed generate $P_D(Y,X,A)$. We first define $P(y_1\vert\, x, a)$ as follows\\
        \[  
            P(y_1\vert\, x, a) := 
            \begin{cases}
                \dfrac{P_D(y_1\vert\, x, a) - P(c_1\vert\, y_0, a)}{P(c_0\vert\, y_1, a)-P(c_1\vert\, y_0,  a)}
                & \text{if }P(c_0\vert\, y_1, a)\neq P(c_1\vert\, y_0, a)\\
                % &\\
                P(y_1)& \text{otherwise}
            \end{cases}
        \]\\
        It's clear that $0\leq P(y\vert\, x)\leq 1$ since $P_D(y_1\vert\, x, a_1) \in [P(c_1\vert\, y_0,  a_1),P(c_0\vert\, y_1, a_1)]$ if $P(c_1\vert\, y_0,  a_1)<P(c_0\vert\, y_1, a_1)$, or $P_D(y_1\vert\, x, a_1) \in [P(c_0\vert\, y_1, a_1), P(c_1\vert\, y_0,  a_1)]$ otherwise. But it's also clear that 
        $\EE_{x\sim P_D}\left[ P(y_1\vert\, x,a)\vert\, A \right]= P(y_1)$ as it follows directly from the first condition that
        \[
            P(y_1) = \frac{P_D(y_1\vert\, a) - P(c_1\vert\, y_0, a)}{P(c_0\vert\, y_1, a)-P(c_1\vert\, y_0,  a)}
        \]
        To show that a solution exists let's first notice that if $P(y_1)$ belongs in the interval (\ref{eq:interval}) then we have
        \begin{align*}
            \max_{x\in X} P_D(y_1\vert\, x,a)P(y_1)\leq P_D(y_1\vert\, a)\leq P(y_1)+P(y_0)\min_{x\in X} P_D(y_1\vert\, x,a)
        \end{align*}
        By Bolzano’s theorem we have that a suitable choice of $P(c\vert\, y, a)$ must exists.
    \end{proof}
    \end{theorem}
\begin{example}\label{ex: label+parity}
Consider the following datasets:\\
\begin{minipage}[l]{0.49\linewidth}
\centering
\[
    \begin{array}{c|c|c|c|c|}
        \# \text{ of} &\ Y =  \ &\  X= \ &\  A= \ \\
         \text{copies} &\ \{+,-\} \ &\  \text{`degree'} \ &\  \text{`sex'} \ \\
        \hline
        \hline
        \times10  & - & \text{`b.sc'} & \male\\\hline
        \times8   & - & \text{`m.sc'} & \male\\\hline
        \times24  & + & \text{`m.sc'} & \male\\\hline
        \times21  & + & \text{`ph.d'} &\male\\\hline\hline
        \times20  & - & \text{`b.sc'} & \female\\\hline
        \times4   & - & \text{`m.sc'} & \female\\\hline
        \times4   & + & \text{`m.sc'} & \female\\\hline
        \times3   & - & \text{`ph.d'} & \female\\\hline
        \times6   & + & \text{`ph.d'} &\female\\\hline
        \hline
    \end{array}
\]
\vspace{0.22\linewidth}
Toy Dataset $D_a$
\vspace{0.1\linewidth}
\end{minipage}
\begin{minipage}[r]{0.49\linewidth}
\centering
\[
    \begin{array}{c|c|c|c|c|}
        \# \text{ of} &\ Y =  \ &\  X= \ &\  A= \ \\
         \text{copies} &\ \{+,-\} \ &\  \text{`degree'} \ &\  \text{`sex'} \ \\
        \hline
        \hline
        \times10 & + & \text{`b.sc'} & \male\\\hline
        \times20 & - & \text{`b.sc'} & \male\\\hline
        \times20 & + & \text{`m.sc'} & \male\\\hline
        \times10 & - & \text{`m.sc'} & \male\\\hline
        \times20 & + & \text{`ph.d'} & \male\\\hline
        \times10 & - & \text{`ph.d'} & \male\\\hline\hline
        \times10 & + & \text{`b.sc'} & \female\\\hline
        \times20 & - & \text{`b.sc'} & \female\\\hline
        \times10 & + & \text{`m.sc'} & \female\\\hline
        \times20 & - & \text{`m.sc'} & \female\\\hline
        \times20 & + & \text{`ph.d'} & \female\\\hline
        \times10 & - & \text{`ph.d'} & \female\\\hline
        \hline
    \end{array}
\]
Toy Dataset $D_b$
\vspace{0.1\linewidth}
\end{minipage}
Despite not knowing how well the datasets represent the probabilities from which they were sampled, using Theorem \ref{Teo:stat+label} we can see why it is improbable for dataset $D_a$ to be the result of a fair distribution (according to the statistical parity notion) where label bias has been introduced. For male candidates we have 
\begin{align*}
    &\max_{x\in X} P_D (y_0\vert\, x, a) = P(Y=-\vert\, X=\text{`b.sc'},A=\male)\approx 1\\*
    &\max_{x\in X} P_D (y_1\vert\, x, a) = P(Y=+\vert\, X=\text{`ph.d'},A=\male)\approx 1\\*
    &P_D (y_1\vert\, a) = P(Y=+\vert\, A=\male)\approx 5/7
\end{align*}
while for female candidates we have
\begin{align*}
    &\max_{x\in X} P_D (y_0\vert\, x, a)= P(Y=-\vert\, X=\text{`b.sc'},A=\female)\approx 1\\*
    &\max_{x\in X} P_D (y_1\vert\, x, a) = P(Y=+\vert\, X=\text{`ph.d'},A=\female)\approx 2/3\\*
    &P_D (y_1\vert\, a) = P(Y=+\vert\, A=\female)\approx 10/37
\end{align*}
So, substituting in Equation (\ref{eq:interval}), we get $\{5/7\}\cap [10/37, 15/37] = \emptyset$.
On the other hand, similar calculations for dataset $D_b$ indicate that it might be the outcome of label bias, since $[1/3,5/6]\cap [1/6,2/3] = [1/3,2/3]$. Theorem \ref{eq:interval} then guarantees the existence of label-biasing distributions $P(c\, \vert\, y,a)$ capable of generating the dataset.
\end{example}
\subsection{Statistical Parity and Selection Bias}
The results presented in this section prove a somewhat counter-intuitive result: even if a distribution that satisfies statistical parity undergoes selection bias and we are able to retrieve the original fair distribution, the demographic disparity difference might (and under stricter conditions, must) still remain strictly positive nonetheless.
We start by showing, similar to the previous case, the set of possible $P(k\vert, y,a)$ that may satisfy the model.
\begin{theorem}\label{teo:sampling+parity}
    The conditional distributions $P(k\,\vert\,  a,y)$ that satisfy the following conditions
    \begin{align*}
        \frac{P_D(y_1\vert\, a_1)}{P_D(y_0\vert\, a_1)}  \frac{P_D(y_0\vert\, a_0)}{P_D(y_1\vert\,  a_0)} = \frac{P(k_1\vert\, y_1,a_1)}{P(k_1\vert\, y_0,a_1)} \frac{P(k_1\vert\, y_0,a_0)}{P(k_1\vert\, y_1,a_0)}
    \end{align*}
    are all and only the distributions that generate $P_D(Y,X,A)$ according to the selection bias model under statistical parity. The set of such distributions is never empty.
    \begin{proof}
    Following Theorem \ref{teo:sampling+unaware} we have \\
        % \begin{multline*}
        % \frac{P(k_1\mid y_1,a_0)}{P(k_1\mid y_0,a_0)}\frac{P_D(y_1\mid  a_1)}{P_D(y_0\mid a_1)} 
        % =  \\
        % \frac{P(k_1\mid y_1,a_0)}{P(k_1\mid y_0,a_0)} \frac{P(y_1)}{P(y_0)} \frac{P(k_1\mid y_1,a_1)}{P(k_1\mid y_0,a_1)} 
        % = \\
        % \frac{P_D(y_1\mid  a_0)}{P_D(y_0\mid a_0)} \frac{P(k_1\mid y_1,a_1)}{P(k_1\mid y_0,a_1)}
        % \end{multline*}
        \[
        \delta_{a_0}\cdot\frac{P_D(y_1\mid  a_1)}{P_D(y_0\mid a_1)} 
        =  
        \delta_{a_0}\cdot\frac{P(y_1)}{P(y_0)} \cdot\delta_{a_1} 
        = 
        \frac{P_D(y_1\mid  a_0)}{P_D(y_0\mid a_0)}\cdot \delta_{a_1} 
        \]\\
    from which the condition follow since $\delta_{a}= P(k_1\mid y_1,a)/P(k_1\mid y_0,a) $. On the other hand, given $P(k_1\vert\, y,a)$ satisfying the equation (which is clearly always possible to find), we can generate $P(y\vert\, x,a)$ by solving the following equation\\
    \[
    \frac{P(y_1\mid x, a)}{P(y_0\mid x, a)}
         :=  \frac{P(k_1\mid y_0,a)}{P(k_1\mid y_1,a)} \frac{P_D(y_1\mid x, a)}{P_D(y_0\mid x, a)}
    \]
    \end{proof}
\end{theorem}
The following theorem shows how the conditional probability of $Y$ given $A$ changes when selection bias happens.
\begin{theorem} \label{teo: stat+sampling}
    Let $P(k_1\vert\, a,y)$ be a conditional selection-biasing distribution. Then the following holds
    \begin{align*}
        \int_{x\in X} P(y_1\vert\, x,a) P_D(x\vert\, a)dx \geq P(y_1\vert\, a)\iff P(k_1 \vert\, y_1,a)\geq P(k_1 \vert\, y_0,a)
    \end{align*}
    Also, all the inequalities are strict unless $Y\indep X\vert\, A=a$
    \begin{proof}
        To show this we must first define the following parametric function 
        \[
            f_\delta(t): = \frac{t}{\delta+(1-\delta)t}
        \]
        Using this function we can rewrite Theorem \ref{teo:sampling+unaware} by saying that the following equations hold
        \begin{align*}
            P(y_1\vert\, x,a) &= f_{\delta_a}(P_D(y_1\vert\, x, a))\\
            P(y_1\vert\,  a) &= f_{\delta_a}(P_D(y_1\vert\,  a))
        \end{align*}\\
        where $\delta_a:={P(k_1\vert\, y_0,a)}/{P(k_1\vert\, y_1,a)}$.
        On the interval $[0,1]$ the function $f_\delta$ is increasing for $\delta\geq 0$ and it's convex when $\delta\in [1,+\infty]$, concave otherwise if $\delta\in [0,1]$. Therefore if $P(k_1 \vert\, a,y_1)\geq P(k_1 \vert\, a,y_0)$ then $\delta_a\in [1,+\infty]$ so, by applying Jensen's inequality
        \begin{gather*}
            \int_{x\in X} P(y_1\vert\, x,a) P_D(x\vert\, a)dx =  
            \int_{x\in X} f_{\delta_a}\bigg(P_D(y_1\vert\,  x, a)\bigg) P_D( x\vert\, A=a)dx \geq \\
            f_{\delta_a}\bigg(\int_{x\in X} P_D(y_1\vert\, x, a) P_D( x\vert\, a)dx \bigg) = 
            f_{\delta_a }\bigg( P_D(y_1\vert\, a)\bigg) =  
            P(y_1\vert\, a)
        \end{gather*}
        and similarly we have the result when $P(k_1 \vert\, a,y_1)\leq P(k_1 \vert\, a,y_0)$.
        Notice that Jensen's inequality is strict when the inequality on $P(k_1 \vert\, a,y)$ is strict unless $Y\indep X\vert\, A=a$.
    \end{proof}
    \end{theorem}
    A direct consequence of this is that if one community is negatively (or positively) impacted by selection bias, while the other is not, the demographic parity difference of the fair model does \textbf{not} converge to zero, even if the original distribution satisfied statistical parity.
    \begin{theorem}
    \label{teo:counteintuitive}
        Let $P$ be fair according to statistical parity.
        Suppose $P(k_1 \vert\, a_0,y_1)\leq P(k_1 \vert\, a_0,y_0)$ and $P(k_1 \vert\, a_1,y_1)\geq P(k_1 \vert\, a_1,y_0)$ with at least one of the inequalities strict. Then the calculated disparity of the fair $P(y_1\vert\, x,a)$ does not converge to zero unless $Y\indep X\vert\, A=a$ for all $a\in A$ where the inequality is strict.
        \begin{proof}
        WLOG let suppose $P(k_1 \vert\, y_1, a_0) < P(k_1 \vert\, y_0, a_0)$.
        As a corollary of the previous theorem we get that\\
            \[
                \int_{x\in X} P(y_1\vert\, x,a_0) P_D(x\vert\, a_0)dx < P(y_1)\leq \int_{x\in X} P(y_1\vert\, x,a_1) P_D(x\vert\, a_1)dx 
            \]\\
            unless $Y\indep X\vert\, A=a_0$. Since\\
            \[
            \sum_{(y,x,a)\in D }\frac
    {P(y_1\mid x,a_i)\mathbb{1}_{a_i}(a)}
    {\vert\{(y,x,a)\in D\colon a = a_i \} \vert}\xrightarrow{\lvert D \rvert \to \infty} \int_{x\in X} P(y_1\vert\, x,a_i) P_D(x\vert\, a_i)dx
            \]\\
            the DPD doesn't converge to zero.
        \end{proof}
    \end{theorem}
Even though we have stated this theorem for $P(y_1\vert x, a)$, the result applies to any model that aims to address selection bias, regardless of whether it is the original model or not. As long as we believe that selection bias was detrimental (or beneficial) to one community but not the other, the DPD of \emph{any} distribution that used to satisfy statistical parity will not converge to zero. Additionally, even though the condition that the two populations have been treated differently is necessary to prove the behavior of the demographic parity difference, in general, Theorem \ref{teo: stat+sampling} shows that whenever selection bias occurs, there are no theoretical guarantees that the demographic parity difference will ever approach zero, even if both populations were affected in the same way. {When bias is introduced into a distribution some of the measures become biased, this can happen to the bias measures themselves}.

\subsection{WAE and Label Bias}\label{subsec:wae+label}
In the next sections, we will assume that our fair-world definition follows the WAE principle and the distribution $P(y,x,a)$ is fair since $Y\indep A\vert X$.
Similarly to Section \ref{subsec:parity+label}, we will start by showing what conditions must $P_D(y\vert x,a)$ follow in order to be generated by that worldview when label bias occurs.
\begin{theorem}\label{teo:disc+unaware}
    Let $P_D$ be a probability distribution resulting from label biasing a distribution satisfying WAE. Then $P_D(y_1\vert\, x, a_0)$ and $P_D(y_1\vert\, x,  a_1)$ are linearly related, that is:
    \[
    \alpha P_D(y_1\vert\, x,a_0) + \beta P_D(y_1\vert\, x, a_1)+\gamma = 0 \quad \text{for some }(\alpha,\beta ,\gamma) \in \RR^3\smallsetminus\{(0,0,0)\}
    \]
    \begin{proof}
    Following Theorem \ref{teo:label bias}, we have this system of linear equations
        \[
        \begin{cases}
            \alpha_0 P(y_1\vert\, x, a_0) + \beta_0 P_D(y_1\vert\, x, a_0)+\gamma_0 = 0\\
            \alpha_1 P(y_1\vert\, x, a_1) + \beta_1 P_D(y_1\vert\, x, a_1)+\gamma_1 = 0\\
            P(y_1\vert\, x, a_0)-P(y_1\vert\, x, a_1)=0
        \end{cases}
        \]
        from which the required relation follows easily.
    \end{proof}
\end{theorem}
As for the case with statistical parity, we can notice that this last relation can be theoretically observed from the distribution we have access to, because it only involves $P_D$. The next proposition shows that if the condition holds then we can find the set of $P(c\vert y,a)$ that generates the distribution
\begin{theorem}\label{teo: all_dist_unaware}
    Suppose the following relation holds
    \[
        \alpha P_D(y_1\vert\, x,a_0) + \beta P_D(y_1\vert\, x, a_1)+\gamma = 0 \quad \text{for some }(\alpha,\beta ,\gamma) \in \RR^3\smallsetminus\{(0,0,0)\}
    \]
    Then the conditional distributions $P(c\vert\, a,y)$ that satisfy the following conditions
    \begin{align*}
        \alpha P(c_1\vert\, y_0, a_0) +& \beta P(c_1\vert\, y_0, a_1)+\gamma = 0\\
        \alpha P(c_0\vert\, y_1, a_0) +& \beta P(c_0\vert\, y_1, a_1)+\gamma = 0\\
        P_D(y_1\vert\, x,a) \in& [P(c_1\vert\, y_0,  a),P(c_0\vert\, y_1, a)]\cup [P(c_0\vert\, y_1,  a),P(c_1\vert\, y_0, a)]
    \end{align*}
    are all and only the distributions that generate $P_D(Y,X,A)$ according to the WAE worldview under label bias. In particular the set of possible $P(c\vert  a,y)$ is not empty.
    \begin{proof}
    WLOG we assume $\alpha\neq 0$ so we can rewrite the equation as
    \[
        P_D(y_1\vert\, x,a_0) = m P_D(y_1\vert\, x,  a_1)+q
    \]
    and the set of conditions as
    \begin{align*}
        P(c_1\vert\, y_0, a_0) &= m P(c_1\vert\, y_0, a_1)+q\\
        P(c_0\vert\, y_1, a_0) &= m P(c_0\vert\, y_1, a_1)+q\\
        P_D(y_1\vert\, x,a) &\in [P(c_1\vert\, y_0,  a),P(c_0\vert\, y_1, a)]\cup [P(c_0\vert\, y_1,  a),P(c_1\vert\, y_0, a)]
    \end{align*}
    where $m,q$ are uniquely defined.
    First let's show that we can find $P(c\vert\, y,a)$ satisfying the aforementioned conditions and with that we can generate $P_D(y\vert\, x,a)$ according to the label bias model. To show the existence of $P(c\vert\, y,a)$ let's notice that we can define
    \[
        P(c_0\vert\, y,a):=\max_{x\in X} P_D(y\vert\, x, a)
    \]
    since $P_D(y\vert\, x, a)$ follows the initial constraint then also $P(c\vert\, y,a)$ has to.
    Now, having $P(c\vert\, y,a)$, we can define $P(y\vert\, x)$ as follow\\
    \[
        P(y_1\vert\, x) :=
        \begin{cases}
                \dfrac{P_D(y_1\vert\, x, a_1) - P(c_1\vert\, y_0, a_1)}{P(c_0\vert\, y_1, a_1)-P(c_1\vert\, y_0,  a_1)}
                & \text{if }P(c_0\vert\, y_1, a_1)\neq P(c_1\vert\, y_0, a_1)\\
                1& \text{otherwise}
        \end{cases} 
    \]\\
    For the same argument in Theorem \ref{Teo:stat+label} we have $0\leq P(y\vert\, x)\leq 1$ so we only need to prove that
    \[
        P_D(y_1\vert\, x,a) = P(c_0\vert\, y_1,a)P(y_1\vert\, x)+ P(c_1\vert\, y_0,a)P(y_0\vert\, x)
    \]
    It's easy to see that the case for $A=a_0$ follows from $A=a_1$ because\\
    \begin{gather*}
        P_D(y_1\vert\, x,a_0) = P(c_0\vert\, y_1,a_0)P(y_1\vert\, x)+ P(c_1\vert\, y_0,a_0)P(y_0\vert\, x)\\
        \iff
        \\
        mP_D(y_1\vert\, x,a_1)+q = [mP(c_0\vert\, y_1,a_1)+q]P(y_1\vert\, x)+ [mP(c_1\vert\, y_0,a_1)+q]P(y_0\vert\, x)\\
        \iff\\
        mP_D(y_1\vert\, x,a_1)= m[P(c_0\vert\, y_1,a_1)P(y_1\vert\, x)+ P(c_1\vert\, y_0,a_1)P(y_0\vert\, x)]
    \end{gather*}\\
    So we need to focus our attention only at the case $A=a_1$.
    If $P(c_0\vert\, y_1, a_1)= P(c_1\vert\, y_0, a_1)$ then $P_D(y_1\vert\, x, a) = P(c_0\vert\, y_1, a)$, therefore\\
    \begin{equation*}
        P_D(y_1\vert\, x, a_1) = P(c_0\vert\, y_1,a_1)P(y_1\vert\, x)+ P(c_1\vert\, y_0,a_1)P(y_0\vert\, x) 
    \end{equation*}\\
    If instead $P(c_0\vert\, y_1, a_1 )\neq P(c_1\vert\, y_0, a_1)$ by substituting $P(y_1\vert\, x)$  we get\\
    \begin{multline*}
        P(c_0\vert  y_1,a_1)P(y_1\vert  x)+ P(c_1\vert  y_0,a_1) P(y_0\vert  x)
        =\\\\
        P(c_0\vert  y_1,a_1) \frac{P_D(y_1\vert  x, a_1) - P(c_1\vert  y_0, a_1)}{P(c_0\vert  y_1, a_1)-P(c_1\vert  y_0,  a_1)} +
        P(c_1\vert  y_0,a_1) \frac{P(c_0\vert  y_1, a_1)-P_D(y_1\vert  x, a_1)}{P(c_0\vert  y_1, a_1)-P(c_1\vert  y_0,  a_1)}
        \\\\
        =P_D(y_1\vert\, x, a_1) \frac{P(c_0\vert\, y_1, a_1)-P(c_1\vert\, y_0,  a_1)}{P(c_0\vert\, y_1, a_1)-P(c_1\vert\, y_0,  a_1)}  = P_D(y_1\vert\, x, a_1) 
    \end{multline*}
    as we wanted.\\\\
    Let's now prove that if $P_D(y_1\vert\, x, a)$ is a result of label bias in a WAE world we get the conditions showed. The condition on $P_D(y_1\vert\, x,a)$ follows from the fact that if\\
    \[
        P_D(y_1\vert\, x,a) = P(c_0\vert\, y_1,a)P(y_1\vert\, x)+ P(c_1\vert\, y_0,a)P(y_0\vert\, x)
    \]\\
    then $P_D(y_1\vert\, x,a)$ is an average of $P(c_0\vert\, y_1,a)$ and $P(c_1\vert\, y_1,a)$.
    For the other conditions we can consider the following linear system\\
    \[
    \begin{cases}
        P_D(y_1\vert\, x,a_0) =  P(c_0\vert\, y_1,a_0)P(y_1\vert\, x)+  P(c_1\vert\, y_0,a_0)P(y_0\vert\, x)
        \\
        P_D(y_1\vert\, x,a_1) =  P(c_0\vert\, y_1,a_1)P(y_1\vert\, x)+  P(c_1\vert\, y_0,a_1)P(y_0\vert\, x)
        \\
        1 =   P(y_1\vert\, x)+   P(y_0\vert\, x)
    \end{cases}
    \]\\
    since $P(y_1\vert\, x)$ is a solution, by Rouché–Capelli theorem ,it follows that\\
    \[
    \det
    \begin{bmatrix}
        P_D(y_1\vert\, x,a_0) & P(c_0\vert\, y_1,a_0)&  P(c_1\vert\, y_0,a_0)
        \\
        P_D(y_1\vert\, x,a_1) &  P(c_0\vert\, y_1,a_1) &  P(c_1\vert\, y_0,a_1) 
        \\
        1 &   1&   1
    \end{bmatrix}\neq 0
    \]\\
    and this is equivalent to the conditions we were trying to prove.
    \end{proof}
\end{theorem}
\begin{example}\label{ex: label+wae}
Let's once again consider the datasets from Example \ref{ex: label+parity}.
This time, the roles played by the datasets are swapped: the first dataset is the one that can be generated by the combination of label bias and the WAE fairness notion, while the other dataset is the one that cannot. In fact, if we calculate the conditional probabilities for the first dataset we obtain:
% \begin{align*}
%     &P(Y = +\vert\ X=\text{`b.sc'}, A=\male) \approx 0\\
%     &P(Y = +\vert\ X=\text{`m.sc'}, A=\male) \approx 3/4\\
%     &P(Y = +\vert\ X=\text{`ph.d'}, A=\male)  \approx 1
% \end{align*}
% for males; and for females:
% \begin{align*}
%     &P(Y = +\vert\ X=\text{`b.sc'}, A=\female)  \approx 0\\
%     &P(Y = +\vert\ X=\text{`m.sc'}, A=\female)  \approx 1/2\\
%    &P(Y = +\vert\ X=\text{`ph.d'}, A=\female)  \approx 2/3
% \end{align*}
\begin{align*}
    &P(Y = +\vert\ X=\text{`b.sc'}, A=\male) \approx 0 &P(Y = +\vert\ X=\text{`b.sc'}, A=\female)  \approx 0\\
    &P(Y = +\vert\ X=\text{`m.sc'}, A=\male) \approx 3/4 &P(Y = +\vert\ X=\text{`m.sc'}, A=\female)  \approx 1/2\\
    &P(Y = +\vert\ X=\text{`ph.d'}, A=\male)  \approx 1 &P(Y = +\vert\ X=\text{`ph.d'}, A=\female)  \approx 2/3
\end{align*}
So the probabilities appear to satisfy 
\[
    P(Y = +\vert\ X=x, A=\female) = \frac{2}{3} P(Y = +\vert\ X=x, A=\male)
\]
and Theorem \ref{teo:disc+unaware} as well.
The other dataset, however, does not satisfy it since
\begin{align*}
    &P(Y = +\vert\ X=\text{`b.sc'}, A=\male)= P(Y = +\vert\ X=\text{`b.sc'}, A=\female)  \approx 1/3\\
    &P(Y = +\vert\ X=\text{`m.sc'}, A=\male)\neq P(Y = +\vert\ X=\text{`m.sc'}, A=\female) \\
    &P(Y = +\vert\ X=\text{`ph.d'}, A=\male) = P(Y = +\vert\ X=\text{`ph.d'}, A=\female) \approx 2/3
\end{align*}
\end{example}
\subsection{WAE and Selection Bias}
The study of this combination of worldview and bias is more straightforward than the previous case: we start, as in the previous cases, by checking whether it is possible to observe this combination in our biased dataset.
\begin{theorem} \label{teo: wae+sampling}
    Let $P_D$ be a probability distribution resulting from selection biasing a distribution satisfying WAE then 
    \[
    \frac{P_D(y_1\vert\, x, a_1)}{P_D(y_0\vert\, x, a_1)} = \alpha\cdot
    \frac{P_D(y_1\vert\, x, a_0)}{P_D(y_0\vert\, x, a_0)}  \text{ for some }\alpha\in [0,+\infty] 
    \]
    \begin{proof}
        It's a direct consequence of Theorem \ref{teo:sampling+unaware}, since $P(y\vert x,a) =P(y\vert x)$ we have\\
        \[
        \frac{P_D(y_1\mid x, a_1)}{P_D(y_0\mid x, a_1)} =  \frac{P(k_1\mid y_1,a_1)}{P(k_1\mid y_0,a_1)} \frac{P(k_1\mid y_0,a_0)}{P(k_1\mid y_1,a_0)}\frac{P(y_1\mid x, a_0)}{P(y_0\mid x, a_0)} 
        \]\\
        so 
        \[
        \alpha = \frac{P(k_1\mid y_1,a_1)}{P(k_1\mid y_0,a_1)} \frac{P(k_1\mid y_0,a_0)}{P(k_1\mid y_1,a_0)}
        \]
    \end{proof}
\end{theorem}
The theorem also already shows what are the possible values of $P(k\vert y,a)$, but we formally state it in the following corollary.
\begin{corollary}\label{teo:sampling+unawareness_dist}
    Suppose the following relation holds
    \[
        \frac{P_D(y_1\vert\, x, a_1)}{P_D(y_0\vert\, x, a_1)} = \alpha\cdot \frac{P_D(y_1\vert\, x, a_0)}{P_D(y_0\vert\, x, a_0)} 
    \]
    Then the conditional distributions $P(k \vert\, a,y)$ that satisfy the following condition
    \begin{align*}
        \frac{P(k_1\vert\, y_1,a_1)}{P(k_1\vert\, y_0,a_1)}  = \alpha\cdot \frac{P(k_1\vert\, y_1,a_0)}{P(k_1\vert\, y_0,a_0)}
    \end{align*}
    are all and only the distributions that generate $P_D(Y,X,A)$ according to the WAE worldview under selection bias.
    \begin{proof}
        Same of Theorem \ref{teo: wae+sampling}.
    \end{proof}
\end{corollary}
\begin{example}
    Since the results of this section are quite similar to the results for Section \ref{subsec:wae+label}, we can utilize a modified version of Example \ref{ex: label+wae}:\\
\begin{minipage}[l]{0.49\linewidth}
\centering
\[
    \begin{array}{c|c|c|c|c|}
        \# \text{ of} &\ Y =  \ &\  X= \ &\  A= \ \\
         \text{copies} &\ \{+,-\} \ &\  \text{`degree'} \ &\  \text{`sex'} \ \\
        \hline
        \hline
        \times10  & - & \text{`b.sc'} & \male\\\hline
        \times40   & - & \text{`m.sc'} & \male\\\hline
        \times30  & + & \text{`m.sc'} & \male\\\hline
        \times10  & - & \text{`ph.d'} & \male\\\hline
        \times10  & + & \text{`ph.d'} &\male\\\hline\hline
        \times20  & - & \text{`b.sc'} & \female\\\hline
        \times20   & - & \text{`m.sc'} & \female\\\hline
        \times10   & + & \text{`m.sc'} & \female\\\hline
        \times30   & - & \text{`ph.d'} & \female\\\hline
        \times20   & + & \text{`ph.d'} &\female\\\hline
        \hline
    \end{array}
\]
\vspace{0.1\linewidth}
Toy Dataset $D_{a'}$
\vspace{0.145\linewidth}
\end{minipage}
\begin{minipage}[r]{0.49\linewidth}
\centering
\[
    \begin{array}{c|c|c|c|c|}
        \# \text{ of} &\ Y =  \ &\  X= \ &\  A= \ \\
         \text{copies} &\ \{+,-\} \ &\  \text{`degree'} \ &\  \text{`sex'} \ \\
        \hline
        \hline
        \times10 & + & \text{`b.sc'} & \male\\\hline
        \times20 & - & \text{`b.sc'} & \male\\\hline
        \times20 & + & \text{`m.sc'} & \male\\\hline
        \times10 & - & \text{`m.sc'} & \male\\\hline
        \times20 & + & \text{`ph.d'} & \male\\\hline
        \times10 & - & \text{`ph.d'} & \male\\\hline\hline
        \times10 & + & \text{`b.sc'} & \female\\\hline
        \times20 & - & \text{`b.sc'} & \female\\\hline
        \times10 & + & \text{`m.sc'} & \female\\\hline
        \times20 & - & \text{`m.sc'} & \female\\\hline
        \times20 & + & \text{`ph.d'} & \female\\\hline
        \times10 & - & \text{`ph.d'} & \female\\\hline
        \hline
    \end{array}
\]
\vspace{0.1\linewidth}
Toy Dataset $D_{b'}$
\end{minipage}
These new datasets have been generated such that the conditioned odds of $(Y=+)$ in $D_a,D_b$ were equal to $P(Y = +\vert X, A)$ in $D_a,D_b$ respectfully. As a consequence in $D_{a'}$ we have:
\begin{align*}
    &P(Y = +\vert\ X=\text{`b.sc'}, A=\male)/P(Y = -\vert\ X=\text{`b.sc'}, A=\male) \approx 0\\
    &P(Y = +\vert\ X=\text{`m.sc'}, A=\male)/P(Y = -\vert\ X=\text{`m.sc'}, A=\male)  \approx 3/4\\
    &P(Y = +\vert\ X=\text{`ph.d'}, A=\male)/P(Y = -\vert\ X=\text{`ph.d'}, A=\male)  \approx 1
\end{align*}
for males; and for females:
\begin{align*}
    &P(Y = +\vert\ X=\text{`b.sc'}, A=\female)/P(Y = -\vert\ X=\text{`b.sc'}, A=\female)  \approx 0\\
    &P(Y = +\vert\ X=\text{`m.sc'}, A=\female)/P(Y = -\vert\ X=\text{`m.sc'}, A=\female)  \approx 1/2\\
    &P(Y = +\vert\ X=\text{`ph.d'}, A=\female)/P(Y = -\vert\ X=\text{`ph.d'}, A=\female)  \approx 2/3
\end{align*}
which satisfies Theorem \ref{teo: wae+sampling}, while for similar reasons $D_{b'}$ does not.
\end{example}
\section{Discussion}
\subsection{Significance and Limitations of the Results}\label{subsec:explanation}
To fully comprehend the impact of the results presented in Section \ref{sec:Worldview and Bias}, a thorough discussion is needed. In primis, we want to clarify that these results do not propose a new fair ML model. Theorems \ref{teo: retrive label+DP}, \ref{teo:sampling+parity}, \ref{teo: all_dist_unaware} and Corollary \ref{teo:sampling+unawareness_dist} all demonstrate how to retrieve the original fair probability from the unfair one. In principle, this means that one would be able to update the original scores of a model to new fair scores. But since the conversion of scores is monotonic for each sensitive attribute, any prediction based on these new scores would be equivalent to the selection of different thresholds during prediction. In other words, adjusting the scores to achieve fairness does not change the underlying predictions; it only affects the threshold at which the predictions are made. This observation is also corroborated by results from \cite{corbett2017algorithmic, menon2018cost}, which state that when provided with an unfair probability distribution of the data, the optimal model can be obtained by choosing labels using different thresholds for the sensitive attributes. This indicates that these claims seem to hold even when the Fair World Framework is adopted.

On the other hand, the aforementioned theorems and corollary also demonstrate that the set of possible biasing distributions, when non-empty, contains multiple elements. Therefore, without prior information or assumptions, it is impossible to determine which of the possible thresholds is the most suitable according to the fair distribution.

Of greater importance, in our view, are Theorems \ref{Teo:stat+label}, \ref{teo:disc+unaware}, and \ref{teo: wae+sampling}. These theorems establish the conditions that unfair distributions must meet if they result from a particular combination of bias and worldview. To fully utilize these theorems, complete knowledge of $P_D(y\vert x,a)$ is required. Also, since our mathematical exploration assumes the bias to be conditionally independent from $X$, we acknowledge that these results may not perfectly model the real world. However, the significance of these theorems lies in the insights they provide, even when relaxed. Due to their importance, it is worthwhile to take a moment to discuss each of these theorems individually.
\begin{itemize}
    \item Theorem \ref{Teo:stat+label} and its Corollary \ref{coro: example} are dependent on the value $\max_{x\in X}P_D(y\vert x,a)$, specifically requiring $\max_{x\in X}P_D(y\vert x,a)<1$. Although, in a realistic scenario, it is highly likely to have data points with conditional probabilities close to one, these propositions demonstrate that label bias is associated with certainty of prediction. This implies that if a fairness-unaware model is underperforming on unfair data, exhibiting disparities based on a sensitive attribute, one of the possible reasons for this could be label bias in the dataset. Therefore, it might be possible, based solely on the unfair distribution, to determine not only which fairness measure to employ but also which type of bias to address.\\
    \item Theorem \ref{teo:disc+unaware} requires a linear relationship between conditional probabilities of different sensitive attributes. Given that the WAE worldview assumes the label to be independent from the sensitive attribute, given the rest of the attributes, one effective intervention to address bias would be to remove the sensitive attribute from the dataset before training. This procedure is known as ``Fairness by Unawareness". This procedure is frequently criticized by the scientific community for two significant reasons. Firstly, there may be sufficient information contained within the remaining attributes to predict the sensitive attribute, leading to an effect known as redlining. Consequently, removing the sensitive attribute may prove ineffective, if not entirely futile. Secondly, without the inclusion of the sensitive attribute, achieving and certifying fair-aware models becomes a challenging task, as even conventional fairness measures cannot be computed in a canonical manner. Nevertheless, the proposed theorem provides valuable insights into when ``Fairness by Unawareness" could be a viable procedure. Given a probabilistic model, there should be a high linear correlation between the scores of data points where the sensitive attribute is manually swapped.\\
    \item Theorem \ref{teo: wae+sampling}  shares similar remarks as the previous theorem, with the distinction that the linear correlation should be computed not on the scores themselves, but on their odds. This distinction provides additional information about the type of bias that has occurred.
\end{itemize}
Finally, in our opinion, the most crucial result comes from Theorem \ref{teo: stat+sampling} and Theorem \ref{teo:counteintuitive}. These theorems state that minimizing the DPD (or equivalent measure) when sampling bias has occurred does not lead to an inherently fair distribution. 
This finding has multiple consequences and interpretations. 

The most general one is that imposing fairness constraints without considering the specific type of bias can have negative effects. A practitioner should always take into consideration the type of bias present in their data, and now we have theoretically sound reasons to support this approach.
Another important point that emerges from these results is the intrinsic difference between models based on the timing of fairness optimization: pre-processing methods operate under distinct assumptions compared to in-processing and post-processing ones. More specifically, any in-processing and post-processing method that aims to minimize the disparity of the output data will encounter the effect described by the theorem, as they do not consider selection bias as a potential source of bias. In contrast, pre-processing methods typically focus on minimizing the DPD in the input data. Depending on how they transform the data, pre-processing methods have the potential to address various types of bias. 

A perfect example of the capability of pre-processing methods is the ``reweighing'' algorithm proposed by Kamiran and Calders~\cite{kamiran2012data}. It assigns a weight to each datapoint based on the sensitive attribute and the label. The weights are chosen such that in the now-weighted dataset it holds $P(y, a)= P_D(y)P_D(a)$ for all $a\in A,y\in Y$. With calculations similar to Theorem \ref{teo:sampling+unaware}, it is possible to show that the new probabilities for the dataset follow the relation
\[
    \frac{P(y_1\mid x, a)}{P(y_0\mid x, a)} =  \frac{P_D(a\mid y_0)}{P_D(a\mid y_1)} \frac{P_D(y_1\mid x, a)}{P_D(y_0\mid x, a)}
\]
which is the inverse of the equation for selection bias when $P(y)=P_D(y)$. Therefore, we not only have an example of a pre-processing method that aims to address a different case than label bias but we can also assert that the underlying assumption of reweighing is that a specific instance of sampling bias has occurred.
\subsection{Unlocking Fairness}
\begin{figure}[t!]
\centering
\begin{subfigure}[t]{\linewidth}
\includegraphics[width=0.52\linewidth]{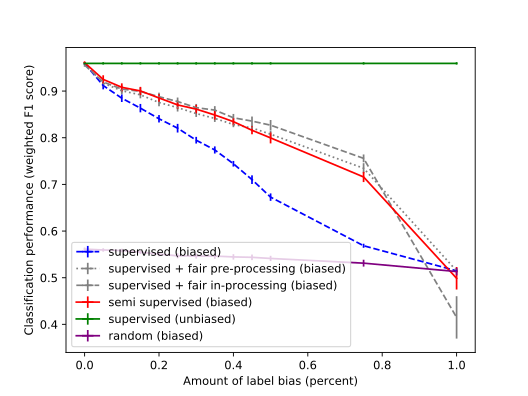}
\includegraphics[width=0.52\linewidth]{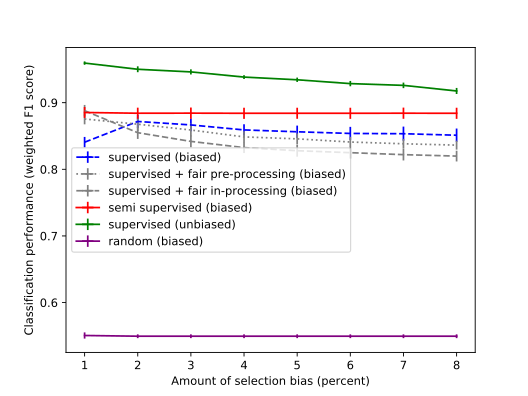}
\caption{F1 score on fair data under label bias (left) and selection bias (right)}
\end{subfigure}
\begin{subfigure}[t]{\linewidth}
\includegraphics[width=0.52\linewidth]{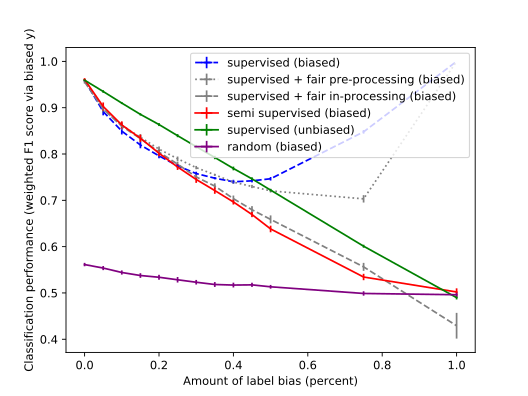}
\includegraphics[width=0.52\linewidth]{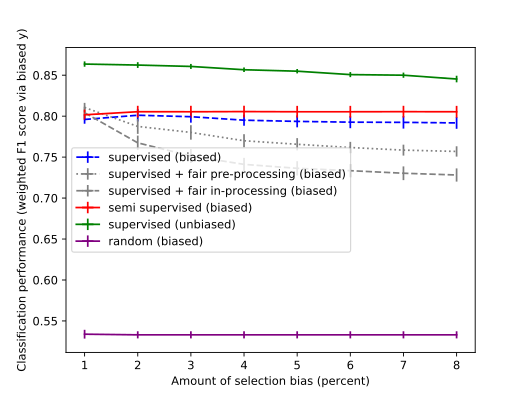}
\caption{F1 score on unfair data under label bias (left) and selection bias (right)}
\end{subfigure}
\begin{subfigure}[t]{\linewidth}
\includegraphics[width=0.52\linewidth]{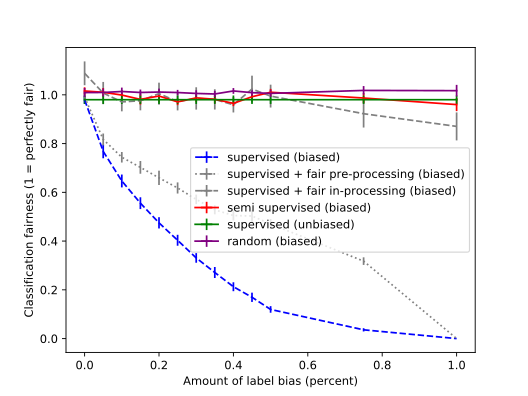}
\includegraphics[width=0.52\linewidth]{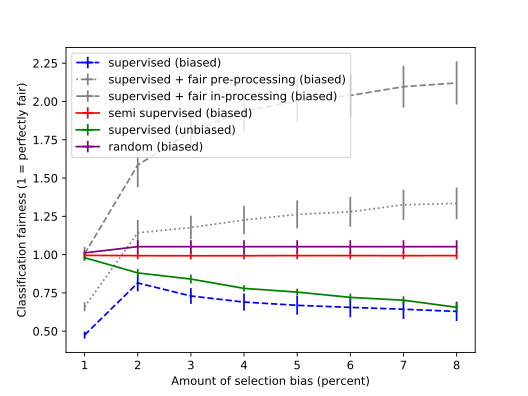}
\caption{Disparity of the models under label bias (left) and selection bias (right)}
% \label{fig:wick_label}
\end{subfigure}
\caption{Results from \cite{wick2019unlocking} on the fairness-accuracy trade-off.}
\label{fig:wick}
\end{figure}
In their paper ``Unlocking fairness: a trade-off revisited''~\cite{wick2019unlocking}, Wick et al. argue that the accuracy-fairness trade-off is a false notion. They show that when the F1-score of a fair model is computed on a fair dataset, fairness and accuracy positively correlate.

To demonstrate this, they generated fair data using statistical parity and WAE as their fairness assumptions. They then introduced label and selection bias into the dataset with increasing intensity. Finally, they evaluated six different models on both a biased and an unbiased test set and evaluated their fairness on the unbiased one.\\
The six models are:
\begin{itemize}
    \item A fairness-agnostic supervised model trained on the biased dataset.
    \item The same model but trained on a pre-processed dataset using the ``reweighing'' method~\cite{kamiran2012data}.
    \item The supervised model where the square of the DPD has been added as an extra term in the cost function to make the model fairness-aware.
    \item The same model but where the extra term has been computed on an unbiased and unlabeled validation set, making it semi-supervised.
    \item The original supervised model but trained on the unbiased dataset as a baseline.
    \item A random model assigning labels according to the biased distribution.
\end{itemize}
Their results are shown in Figure \ref{fig:wick}. The results clearly show what they claim, but they also acknowledge how all the fairness-aware models, besides from the semi-supervised one, ``all succumb to selection bias''. Our theoretical framework provides an explanation for these findings. As stated in Theorem \ref{teo:label bias}, under label bias, $P_D(x,a)=P(x,a)$, so the DPD of a probabilistic model $P_M(y_1\vert\, x,a)$ converges to the same value when evaluated on the biased data or unbiased data. Therefore, when the original distribution satisfies statistical parity, any model that tries to minimize the DPD (regardless of where it is calculated) will indeed approach the fair distribution. However, different results should be expected when considering the case for selection bias: as shown by Theorems \ref{teo: stat+sampling} and \ref{teo:counteintuitive}, under selection bias, the DPD of the fair model evaluated on the biased dataset doesn't converge to its original value. Since both the fairness-aware supervised model and the pre-processed supervised model aim to achieve statistical parity on the biased data, their DPDs approach zero when evaluated on the biased dataset. But the DPD of the fair model is \emph{not} zero, so both models get further and further away from the fair distribution and their fairness scores overshoot the target. On the other hand the semi-supervised model evaluates the DPD on an unbiased dataset, so the DPD for that model does indeed correspond to the fairness behaviour on fair data. Therefore, even if the fair distribution in the experiment for selection bias had a small disparity, the performance and fairness of the semi-supervised model remain constant and unfooled by the bias injection. 

Additionally, we can observe that the performance of the unfair supervised model on biased data tends to degrade as the label bias increases. Then, at some point, it starts to improve again. This observation aligns with what was explained in Section \ref{subsec:explanation}. Label bias is closely related to the certainty of prediction: as label bias increases, so does the level of uncertainty in the predictions. At a certain point, when the bias becomes extremely strong, the model's prediction becomes solely dependent on the sensitive attribute. As the bias increases further, this decision based solely on the sensitive attribute becomes more and more accurate, resulting in an improvement in performance.

Finally, when we examine the model in which reweighing was applied, we can observe that it specifically addresses selection bias while failing to address label bias, as predicted.
\section{Experiments}
\begin{figure}[t!]
\centering
\begin{subfigure}[t]{\linewidth}
\includegraphics[width=0.5\linewidth]{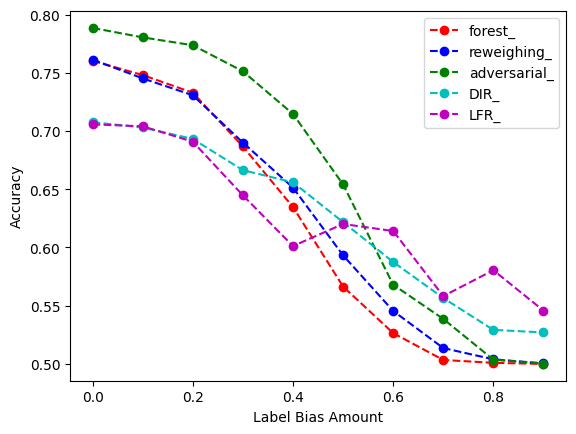}
\includegraphics[width=0.5\linewidth]{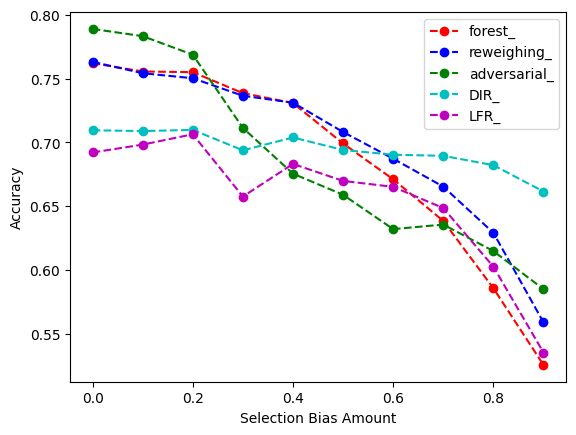}
\caption{Accuracy on fair data under label bias (left) and selection bias (right)}
\end{subfigure}
\begin{subfigure}[t]{\linewidth}
\includegraphics[width=0.5\linewidth]{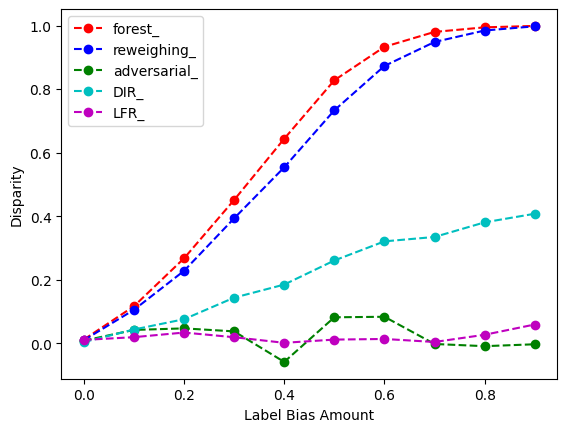}
\includegraphics[width=0.5\linewidth]{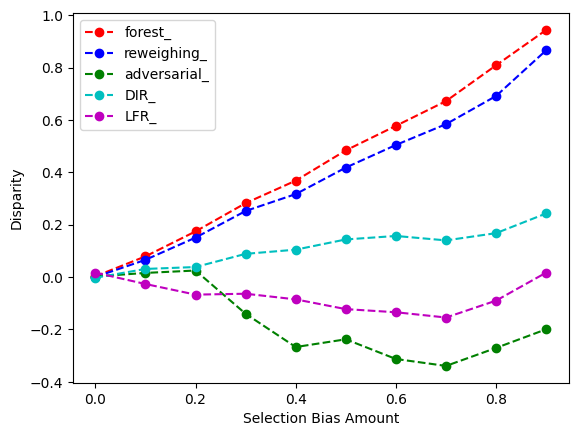}
\caption{Disparity of the models under label bias (left) and selection bias (right)}
% \label{fig:our_results}
\end{subfigure}
\caption{Experimental results}
\label{fig:our_results}
\end{figure}
Similarly to Wick et al., we conducted experiments on artificial data using models from the AIF360 library \cite{bellamy2019ai}. The main difference from the previous data is that in our experiments, we focused on statistical parity as the fairness criterion, without incorporating additional fairness assumptions. This was done to maintain consistency and avoid introducing potential biases in the results. To achieve this, we utilized a Bayesian network. For each datapoint, we independently generated a sensitive attribute and a label. This ensured that statistical parity was maintained. Subsequently, we generated all the other features based on the generated label and sensitive attribute, with each feature being generated independently from the others. Then, we introduced increasing amounts of label and selection bias into the datasets. The experiment was repeated 10 times, and the results were averaged across the repetitions. 
The following algorithms were used in the experiments:
\begin{itemize}
    \item A fairness-agnostic random-forest classifier was used as a baseline model.
    
    \item The same model but trained on a pre-processed dataset using the ``reweighing” method.
    \item The adversarial debiasing algorithm from \cite{adversarial}.
    \item The initial random-forest classifier on a pre-processed dataset using the ``Disparate Impact Remover (DIR)” algorithm \cite{feldman2015certifying}.
    \item The ``Learn Fair Representation (LFR)" algorithm from \cite{zemel2013learning}.
\end{itemize}
All methods were trained on biased data and evaluated on fair data, prior to pre-processing. Results are shown in Figure \ref{fig:our_results}. As for the previous section, we can observe that the adversarial debiasing method, being an in-processing model, effectively minimizes the disparity of the data for label bias. However, it does not perform as well in addressing selection bias, resulting in disproportionate impact on one of the sensitive groups. This is also reflected in the performance of the classifier, as it performs the best overall in mitigating label bias but struggles when it comes to addressing sampling bias.

% Similarly does, although to a lesser degree, the LFR algorithm which is often classified as a pre-processing method but it works similarly to an in-processing one. 
Similarly, the disparity of the LFR algorithm exhibits a behavior comparable to an in-processing method, albeit to a lesser degree. This is not surprising since, although LFR is often classified as a pre-processing method, it functions similarly to an in-processing one.
However, qualitatively, the performance of the algorithm is lacking for both biases. 

Interestingly, the DIR method does not appear to be as significantly affected by the increasing bias compared to the other methods. We believe this could be attributed to the way the fair data are generated. In particular, the fact that each feature is conditionally independent from one another given the label and sensitive attribute aligns closely with some of the assumptions of the DIR algorithm, which also independently modifies the values of each attribute.

Finally, the performance of the reweighing algorithm was not as expected. While the behavior of the model under label bias can be explained, as the algorithm specifically addresses selection bias, the performance of the model on its bias of choice falls short. 
We cannot provide an theoretically sound explanation as why this is the case, but we hypothesize this is a consequence of using a random-forest classifier as the main method, which could be less sensitive than other models to the reweighing method.

\section{Conclusion and Future Work}
Our work shows important connections between fairness conditions and bias injections. Proving theoretical conditions that a distribution must satisfy in order to understand if the original distribution has been biased is of primary importance for the future of the topic of fairness. Having these conditions (like Theorems \ref{Teo:stat+label}, \ref{teo:disc+unaware}, \ref{teo: wae+sampling}) not only hint on which fairness definition should be used but also what the fair distribution should look like. This would relieve practitioners from the burden of blindly choosing and minimizing a fairness metric. On the other hand Theorems \ref{teo: stat+sampling} and \ref{teo:counteintuitive} show the risk of using fairness metrics as a minimization objectives without understanding how bias might influence them: the fair distribution could not be between those that minimize the chosen measure.

Nonetheless much work is still needed: the conditions show in this paper are theoretical in nature and detecting when a distribution might satisfy them still requires work. First of all having perfect access to $P_D(y\vert x,a)$, as many of the theorems require, poses a significant challenge in the real world. Also, even if we had access to the distribution, conditions like those shown in Corollary \ref{coro: example} are unrealistically strict and must be relaxed to be able to work with them.

\section*{Declarations}

\begin{itemize}
\item Funding: The first author is supported by the AXA joint research initiative (CS15893) and the third author is supported by FWO (Flanders).  
\item Conflict of interest: the authors declare no conflict of interest outside their institution.
\item Ethics approval: Not applicable
\item Consent to participate: Not applicable
\item Consent for publication: Not applicable
\item Availability of data and materials: Not applicable
\item Code availability: \url{https://github.com/Up2Iso/label_selection_bias} 
\item Authors' contributions: All authors contributed to the study conception and design. Material preparation, data collection and analysis were performed by the first author. The first draft of the manuscript was written by the first author and all authors commented on previous versions of the manuscript. All authors read and approved the final manuscript.
\end{itemize}

\bibliography{references.bib}
\end{document}